\documentclass{article}

\usepackage{microtype}
\usepackage{graphicx}
\usepackage{booktabs} % for professional tables
\usepackage[utf8]{inputenc}

\usepackage{graphicx}
\usepackage{times}   % assumes new font selection scheme installed
\usepackage{amsmath} % assumes amsmath package installed
\usepackage{amssymb} % assumes amsmath package installed
\usepackage{mathtools}
\usepackage{ntheorem}
\usepackage{bm}
\usepackage{algorithm}
\usepackage{algpseudocode}
\usepackage{subcaption}
\usepackage{epstopdf}
\usepackage{stfloats}

\usepackage{tikz}
\usetikzlibrary{calc}
\tikzset{mynode/.style={draw,circle, minimum size = 0.7cm}}
\usetikzlibrary{positioning,shapes,arrows}
\usepackage{xcolor}
\usepackage{pgfplots}
\usetikzlibrary{external}
\usepackage{hyperref}

\usepackage[accepted]{icml2019}

\icmltitlerunning{AReS and MaRS - Adversarial and MMD-Minimizing Regression for SDEs}

\begin{document}

\twocolumn[
\icmltitle{AReS and MaRS - Adversarial and MMD-Minimizing Regression for SDEs}



\icmlsetsymbol{equal}{*}

\begin{icmlauthorlist}
\icmlauthor{Gabriele Abbati}{equal,Ox}
\icmlauthor{Philippe Wenk}{equal,LAS,CLS}
\icmlauthor{Michael A Osborne}{Ox}
\icmlauthor{Andreas Krause}{LAS}
\icmlauthor{Bernhard Schölkopf}{EI}
\icmlauthor{Stefan Bauer}{EI}
\end{icmlauthorlist}

\icmlaffiliation{Ox}{Department of Engineering Science, University of Oxford}
\icmlaffiliation{LAS}{Learning and Adaptive Systems Group, ETH Zürich}
\icmlaffiliation{CLS}{Max Planck ETH Center for Learning Systems}
\icmlaffiliation{EI}{Empirical Inference Group, Max Planck Institute for Intelligent Systems}

\icmlcorrespondingauthor{Gabriele Abbati}{gabb@robots.ox.ac.uk}
\icmlcorrespondingauthor{Philippe Wenk}{wenkph@ethz.ch}

\icmlkeywords{Gaussian Process, Stochastic Differential Equations, Gradient Matching, Ornstein-Uhlenbeck process, parameter inference}

\vskip 0.3in
]



\printAffiliationsAndNotice{\icmlEqualContribution} % otherwise use the standard text.

\begin{abstract}
Stochastic differential equations are an important modeling class in many disciplines. Consequently, there exist many methods relying on various discretization and numerical integration schemes. In this paper, we propose a novel, probabilistic model for estimating the drift and diffusion given noisy observations of the underlying stochastic system. Using state-of-the-art adversarial and moment matching inference techniques, we avoid the discretization schemes of classical approaches. This leads to significant improvements in parameter accuracy and robustness given random initial guesses. On four established benchmark systems, we compare the performance of our algorithms to state-of-the-art solutions based on extended Kalman filtering and Gaussian processes.
\end{abstract}

\section{Introduction}
% Introduction

Modeling discretely observed time series is a challenging problem that arises in quantitative sciences and many fields of engineering. While it is possible to tackle such problems with difference equations or ordinary differential equations (ODEs), both approaches suffer from serious drawbacks. Difference equations are difficult to apply if the observation times are unevenly distributed. Furthermore, they do not generalize well across observation frequencies, while natural laws do not care about this artificial construct. ODEs can deal with these challenges, but fail to incorporate the inherent stochastic behavior present in many physical, chemical or biological systems. These effects can be captured by introducing stochasticity in the dynamics model, which brings to stochastic differential equations (SDEs). In this paper, we use exclusively the Itô-form
\begin{equation}
	\mathrm{d} \mathbf{x}(t) = \mathbf{f}(\mathbf{x}(t), \bm{\theta}) \mathrm{d}t + \mathbf{g}(\mathbf{x}(t), \bm{\theta}) \mathrm{d} \mathbf{w}(t),
\end{equation}
where $\mathbf{x}(t)$ is the time-dependent vector of states we would like to model, $\bm{\theta}$ collects the parameters of the model, $\mathbf{f}$ is the drift term, $\mathbf{g}$ is the matrix-valued diffusion function and $\mathbf{w}(t)$ is a Wiener-process of the same dimension as the state vector $\mathbf{x}$.

While SDEs can efficiently capture stochasticity and deal with unevenly spaced observation times and frequency, inference is rather challenging. Due to the stochasticity of $\mathbf{w}(t)$, the state vector $\mathbf{x}(t)$ is itself a random variable. Except for few special cases, it is not possible to find an analytic solution for the statistics of $\mathbf{x}(t)$ for general drift and diffusion terms. The problem is even more challenging if we were to condition on or state-estimate some discrete time observations $\mathbf{y}$ (filtering/smoothing) or infer some statistics for the parameters $\bm{\theta}$ (parameter inference). It is well known that the parameter inference problem is a difficult task, with most approaches either being very sensitive to initialization \citep{picchini2007_SDEToolbox}, strongly dependent on the choice of hyperparameters like the spacing of the integration grid \citep{bhat2015parameter} or using excessive amount of computational resources even for small scale systems and state-of-the-art implementation \citep{ryder2018}.

The difficulty of the parameter estimation problem of estimating parameters of drift and diffusion under observational noise is readily exemplified by the fact that even major scientific programming environment providers like $\mathrm{MATLAB}$ are still lack an established toolbox for practical use. In this paper, we will take a step into a novel direction tackling this open and exciting research question.

\subsection{Related Work}

While it is impossible to cover all related research efforts, we would like to give a quick overview by mentioning some of the most relevant. For a more in-depth discussion, we recommend \citet{tronarp2018_Stota}, who provide an excellent review of the current state-of-the-art smoothing schemes. Moreover, \citet{sorensen2004_overview}, \citet{nielsen2000_overview} and \citet{hurn2007seeing} provide extensive explanations of the more traditional approaches.

Most classical methods rely on calculating the probability of sample paths $\bm{x}$ conditioned on the system parameters $\bm{\theta}$, denoted as $p(\mathbf{x} | \bm{\theta})$. Since $p(\mathbf{x} | \bm{\theta})$ is usually analytically intractable, approximation schemes are necessary. \citet{MCMCSTOTAelerian2001} and \citet{MCMCSTOTAeraker2001} use the Euler-Maruyama discretization to approximate $p(\mathbf{x} | \bm{\theta})$ on a fixed, fine grid of artificial observation times later to be leveraged in a MCMC sampling scheme. \citet{MCMCNEWpieschner2018bayesian} and \citet{MCMCNEWvan2017bayesian} subsequently refine this approach with improved bridge constructs and incorporated partial observability. \citet{ryder2018} follow up on this idea by combining discretization procedures with variational inference. \citet{sarkka2015posteriorCOMPARISON} investigate different approximation methods based on Kalman filtering, while \citet{archambeau2007_VGPA} and \citet{vrettas2015variational} use a variational Gaussian process-based approximation. Finally, it should be mentioned that $p(\mathbf{x} | \bm{\theta})$ can be inferred by solving the Fokker-Planck-Kolmogorov equations using standard methods for PDEs \citep{hurn1999estimating, ait2002maximum}.

Instead of approximating $p(\mathbf{x}, \bm{\theta})$ in a variational fashion, Gaussian processes can as well be used to directly model $\mathbf{f}(\mathbf{x}, \bm{\theta})$ and $\mathbf{g}(\mathbf{x}, \bm{\theta})$, ignoring in this way any prior knowledge about their parametric form. This approach was investigated by \citet{ruttor2013approximate}, whose linearization and discretization assumptions which were later relaxed by \citet{yildiz2018learning}. While we will show in our experiments that these methods can be used for parameter estimation if the parametric form of drift and diffusion are known, it should be noted that parameter inference was not the original goal of their work.

\subsection{Our Work}

To the best of our knowledge, there are only very few works that try to circumvent calculating $p(\mathbf{x} | \bm{\theta})$ at all. Our approach is probably most closely related to the ideas presented by \citet{riesinger2016solving}. Our proposal relies on the Doss-Sussman transformation \citep{doss1977liens, sussmann1978gap} to reduce the parameter inference problem to parameter inference in an ensemble of random ordinary differential equations (RODEs). These equations can then be solved path-wise using either standard numerical schemes or using the computationally efficient gradient matching scheme of \citet{gorbach2017scalable} as proposed by \citet{bauer2017efficient}.

The path-wise method by \citet{bauer2017efficient} has natural parallelization properties, but there is still an inherent approximation error due to the Monte Carlo estimation of the expectation over the stochastic element in the RODE. Furthermore, their framework imposes severe linearity restrictions on the functional form of the drift $\mathbf{f}(\mathbf{x}, \bm{\theta})$, while it is unable to estimate the diffusion matrix $\mathbf{g}(\mathbf{x}, \bm{\theta})$.

While we will keep their assumption of a constant diffusion matrix, i.e. $\mathbf{g}(\mathbf{x}, \bm{\theta}) = \mathbf{G}$, our approach gets rid of the linearity assumptions on the drift $\mathbf{f}$. Furthermore, we substitute the Monte Carlo approximation by embedding the SDE into a fully statistical framework, allowing for efficient estimation of both $\mathbf{G}$ and $\bm{\theta}$ using state-of-the-art statistical inference methods.

Despite a constant diffusion assumption might seem restrictive at first, such SDE models are widely used, e.g. in chemical engineering \citep{karimi2018bayesian}, civil engineering \citep{jimenez2008estimation}, pharmacology \citep{donnet2013review} and of course in signal processing, control and econometrics. While we believe that this approach could be extended approximately to systems with general diffusion matrices, we leave this for future work.

The contributions of our framework are the following:
\begin{itemize}
	\item We derive a new statistical framework for diffusion and drift parameter estimation of SDEs using the Doss-Sussmann transformation and Gaussian processes. 
	\item We introduce a grid-free, computationally efficient and robust parameter inference scheme that combines a non-parametric Gaussian process model with adversarial loss functions.
	\item We demonstrate that our method is able to estimate constant but non-diagonal diffusion terms of stochastic differential equations without any functional form assumption on the drift.
	\item We show that our method significantly outperforms the state-of-the-art algorithms for SDEs with multi-modal state posteriors, both in terms of diffusion and drift parameter estimation. 
	\item We share and publish our code to facilitate future research at \url{https://github.com/gabb7/AReS-MaRS}.
\end{itemize}
\section{Background}
% BACKGROUND SECTION

In this section, we formalize our problem and introduce the necessary notation and background drawn from Gaussian process-based gradient matching for ODEs.

\subsection{Problem Setting}
We consider SDEs of the form
\begin{equation}
	\mathrm{d}\mathbf{x}(t) = \mathbf{f}(\mathbf{x}(t), \boldsymbol{\theta}) \mathrm{d}t + \mathbf{G} \mathrm{d}\mathbf{w}(t),
	\label{eq:ProbSetting}
\end{equation}
where $\mathbf{x}(t) = [\mathbf{x}_1(t), \dots, \mathbf{x}_K(t)]^\top$ is the $K$-dimensional state vector at time $t$; $\mathrm{d}\mathbf{w}(t)$ are the increments of a standard Wiener process; $\mathbf{f}$ is an arbitrary, potentially highly nonlinear function whose parametric form is known, save for the unknown parameter vector $\boldsymbol{\theta}$; $\mathbf{G}$ is the unknown but constant diffusion matrix. Without loss of generality, we can assume $\mathbf{G}$ to be a lower-diagonal, positive semi-definite matrix.

The system is observed at $N$ arbitrarily spaced time points $\mathbf{t}=[t_1, \dots, t_N]$, subjected to Gaussian observation noise:
\begin{equation}
	\mathbf{y}(t_n) = \mathbf{x}(t_n) + \mathbf{e}(t_n) \hspace{.5cm} \forall \ n=1,\dots, N,
\end{equation}
where we assume the noise variances to be state-dependent but time-independent, i.e.
\begin{equation}
	p(\mathbf{e}(t_n)) = \prod_{k=1}^K \mathcal{N} \left( \mathbf{e}_k(t_n) \mid 0, \sigma_k \right),
	\label{eq:errorModel}
\end{equation}
for $n=1,\dots, N$ and $k=1,\dots, K$.

\subsection{Deterministic ODE Case}
In the context of Bayesian parameter inference for deterministic ordinary differential equations, \citet{calderhead2009accelerating} identify numerical integration as the main culprit for bad computational performance. Thus, they propose to turn the parameter estimation procedure on its head: instead of calculating $p(\mathbf{y} \mid \boldsymbol{\theta})$ using numerical integration, they extract two probabilistic estimates for the derivatives, one using only the noisy observations $\mathbf{y}$ and one using the differential equations. The main challenge is then to combine these two distributions, such that more information about $\mathbf{y}$ can guide towards better parameter estimates $\boldsymbol{\theta}$. For this purpose, \citet{calderhead2009accelerating} propose a product of experts heuristics that was accepted and reused until recently \citet{wenk2018fast} showed that this heuristic leads to severe theoretical issues. They instead propose an alternative graphical model, forcing equality between the data based and the ODE based model save for a Gaussian distributed slack variable.

In this paper, we use another interpretation of gradient matching, which is aimed at finding parameters $\boldsymbol{\theta}$ such that the two distributions over $\dot{\mathbf{x}}$ match as closely as possible. Acknowledging the fact that standard methods like minimizing the KL divergence are not tractable, we use robust moment matching techniques while solving a much harder problem with $\mathbf{G} \neq \mathbf{0}$. However, it should be clear that our methodology could easily be applied to the special case of deterministic ODEs and thus provides an additional contribution towards parameter estimation for systems of ODEs.

\subsection{Notation}
\label{sec:Notation}
Throughout this paper, bold, capital letters describe matrices. Values of a time-dependent quantities such as the state vector $\mathbf{x}(t)$ can be collected in the matrix $\mathbf{X} = [\mathbf{x}(t_1), \dots, \mathbf{x}(t_N)]$ of dimensions $K \times N$, where the $k$-th row collect the $N$ single-state values at times $\mathbf{t} = [t_1, \dots, t_N]$ for the state $k$.

The matrix $\mathbf{X}$ can be vectorized by concatenating its rows and defining in this way the vector $\mathbf{x} = [\mathbf{x}_1, \dots, \mathbf{x}_K]^\top$. This vector should not be confused with $\mathbf{x}(t)$, which is still a time-dependent vector of dimension $K$. 

As we work with Gaussian processes, it is useful to standardize the state observations by subtracting the mean and dividing by the standard deviation, in a state-wise fashion. We define the vector of the data standard deviation $\bm{\sigma}_{\mathbf{y}} = [\sigma_{\mathbf{y}_1}, \dots, \sigma_{\mathbf{y}_K}]$, and the matrix $\mathbf{S}$ as:
\begin{equation}
	\mathbf{S} = \bm{\sigma}_{\mathbf{y}} \otimes \mathbf{I}_N
\end{equation}
where $\otimes$ indicates the Kronecker product and $\mathbf{I}_N$ is the identity matrix of size $N \times N$. Similarly for the means, we can define the $N \times K$ vector $\bm{\mu}_\mathbf{y}$ that contains the $K$ state-wise means of the observations, each repeated $N$ times. Thus the standardize vector $\tilde{\mathbf{x}}$ can be defined as:
\begin{equation}
	\tilde{\mathbf{x}} = \mathbf{S}^{-1}(\mathbf{x} - \bm{\mu}_\mathbf{y}).
\end{equation}

For the sake of clarity, we omit the normalization in the following sections. It should be noted however that in a finite sample setting, standardization strongly improves the performance of GP regression. In our implementation and all the experiments in section \ref{sec:experiments}, we assume a GP prior on the states standardized using the state-wise mean and standard deviation of the observations $\mathbf{y}$.

For coherence with the current Gaussian process-based gradient matching literature, we follow the notation introduced by \citet{calderhead2009accelerating} and \citet{wenk2018fast} wherever possible.

\section{Methods}
\tikzset{mynode/.style={draw,circle, minimum size = 0.7cm}}
\usetikzlibrary{positioning,shapes,arrows}

% METHODS SECTION

In the deterministic case with $\mathbf{G} = \mathbf{0}$, the GP regression model can be directly applied to the states $\mathbf{x}$. However, if $\mathbf{G} \neq \mathbf{0}$, the derivatives of the states with respect to time $t$ no longer exist due to the contributions of the Wiener process. Thus, performing direct gradient matching on the states is not feasible.

%In the case of a non-zero diffusion matrix $\mathbf{G}$, the necessary hypotheses for the straightforward application of GP regression do not hold, as the Wiener process do not allow guarantees of existence for the derivatives of the states.
%We propose to overcome this issue to the above-cited Doss-Sussmann transformation.

\subsection{Latent States Representation}
We propose to tackle this problem by introducing a latent variable $\mathbf{z}$, defined via the linear coordinate transformation
\begin{equation}
	\mathbf{z}(t) = \mathbf{x}(t) - \mathbf{o}(t),
	\label{eq:coordinateTransformation}
\end{equation}
where $\mathbf{o}(t)$ is the solution of the following SDE:
\begin{equation}
	\mathrm{d}\mathbf{o}(t) = -\mathbf{o}(t) + \mathbf{G} \mathrm{d} \mathbf{w}(t).
	\label{eq:OUSDE}
\end{equation}
Without loss of generality, we set $\mathbf{z}(0) = \mathbf{x}(0)$ and thus $\mathbf{o}(0) = 0$. While in principle the framework supports any initial condition as long as $\mathbf{z}(0) + \mathbf{o}(0) = \mathbf{x}(0)$, the reasons for this choice will become clear in section \ref{sec:OUContributions}.

Using Itô's formula, we obtain the following SDE for $\mathbf{z}$
\begin{equation}
	\mathrm{d}\mathbf{z}(t) = \left\{\mathbf{f}(\mathbf{z}(t) + \mathbf{o}(t), \boldsymbol{\theta}) + \mathbf{o}(t)\right\} \mathrm{d}t
	\label{eq:latentDerivs}
\end{equation}
This means that for a given realization of $\mathbf{o}(t)$, we obtain a differentiable latent state $\mathbf{z}(t)$. In principle, we could sample realizations of $\mathbf{o}(t)$ and solve the corresponding deterministic problems, which is equivalent to approximately marginalizing over $\mathbf{o}(t)$. However, it is actually possible to treat this problem statistically, completely bypassing such marginalization. We do this by creating probabilistic generative models for observations and derivatives analytically. The equations are derived in this section, while the final models are shown in Figure \ref{fig:GenerativeModels}.

\subsection{Generative Model for Observations}
Let us define $\mathbf{e}(t)$ as the Gaussian observation error at time $t$. Using the matrix notation introduced in section \ref{sec:Notation}, we can write
\begin{equation}
	\mathbf{Y} = \mathbf{X} + \mathbf{E} = \mathbf{Z} + \mathbf{O} + \mathbf{E},
	\label{eq:ObsSumUnnormalized}
\end{equation}
where $\mathbf{Z}$ and $\mathbf{O}$ are the matrices corresponding to the lower-case variables introduced in the previous section. In contrast to standard GP regression, we have an additional noise term $\mathbf{O}$, which is the result of the stochastic process described by equation \eqref{eq:OUSDE}. As in standard GP regression, it is possible to recover a closed form Gaussian distribution for each term.

\subsubsection{GP Prior}
We assume a zero-mean Gaussian prior over the latent states ${\mathbf{z}}$, whose covariance matrix is given by a kernel function $k(x,y)$, in turn parameterized by the hyperparameter vector $\bm{\phi}$:
\begin{equation}
	p({\mathbf{z}} \mid \bm{\phi}) = \mathcal{N} \left( \mathbf{{z}} \mid \mathbf{0}, \mathbf{C}_{\bm{\phi}} \right).
	\label{eq:GPPrior}
\end{equation}
We treat all state dimensions as independent, meaning that we put independent GP priors with separate hyperparameters $\boldsymbol{\phi}_k$ on the time evolution of each state. Consequently, $\mathbf{C}_{\bm{\phi}}$ is a block diagonal matrix with $K$ blocks each of dimension $N \times N$. The blocks model the correlation over time introduced by the GP prior.

\subsubsection{Error Model}
In equation \eqref{eq:errorModel}, we assume that observational errors are i.i.d. Gaussians uncorrelated over time. The joint distribution of all errors is thus still a Gaussian distribution, whose covariance $\mathbf{T}$ has only diagonal elements given by the GP likelihood variances $\bm{\sigma} = \{\sigma^2_k\}_{k=1}^K$. More precisely:
\begin{equation}
	\mathbf{T} = \bm{\sigma} \otimes \mathbf{I}_N.
\end{equation}
and
\begin{equation}
	p(\mathbf{e} \mid \boldsymbol{\sigma}) = \mathcal{N} \left( \mathbf{e} \mid \mathbf{0}, \mathbf{T} \right).
	\label{eq:ErrorDensity}
\end{equation}

\subsubsection{Ornstein-Uhlenbeck Process}
\label{sec:OUContributions}
Through the coordinate transformation in equation \eqref{eq:coordinateTransformation}, all stochasticity is captured by the stochastic process $\mathbf{o}(t)$ described by equation \eqref{eq:OUSDE}. Such mathematical construct has a closed-form, Gaussian solution and is called Ornstein-Uhlenbeck process. For the one-dimensional case with zero initial condition and unit diffusion
\begin{equation}
	\mathrm{d} \hat{o}(t) = -\hat{o}(t) + \mathrm{d}w(t),
\end{equation}
we get the following mean and covariance:
\begin{align}
	\mathbb{E}[\hat{o}(t)] &= 0\\
	\textrm{cov}[\hat{o}(t_i), \hat{o}(t_j)] &= \frac{1}{2} e^{- | t_i - t_j |}
	 - \frac{1}{2} e^{-(t_i + t_j)}.
	 \label{eq:ou_cov}
\end{align}
Sampling $\hat{o}(t)$ at the $N$ points $\mathbf{t}=[t_1, \dots, t_N]$ yields the vector $\hat{\mathbf{o}}(\mathbf{t}) = [\hat{o}(t_1), \dots, \hat{o}(t_N)]$, which is Gaussian distributed:
\begin{equation}
	p(\hat{\mathbf{o}}(\mathbf{t})) = \mathcal{N}(\hat{\mathbf{o}}(\mathbf{t}) \mid \mathbf{0}, \bm{\Omega}_{\text{one}}),
\end{equation}
where $[\bm{\Omega}_{\text{one}}]_{ij} = \textrm{cov}[\hat{o}(t_i), \hat{o}(t_j)]$ according to \eqref{eq:ou_cov}. In the case of a $K$-dimensional process with identity diffusion, i.e.
\begin{equation}
	\mathrm{d} \hat{\mathbf{o}}(t) = - \hat{\mathbf{o}} + \mathbf{I}_K \mathrm{d} \mathbf{w}(t),
\end{equation}
we can just treat each state dimension as an independent, one-dimensional OU process. Thus, after sampling $\hat{\mathbf{o}}(t)$ $K$ times at the $N$ time points in $\mathbf{t}$ and unrolling the resulting matrix as described in section \ref{sec:Notation}, we get
\begin{equation}
	p(\hat{\mathbf{o}}) = \mathcal{N}(\hat{\mathbf{o}} \mid \mathbf{0}, \boldsymbol{\Omega}),
\end{equation}
where $\boldsymbol{\Omega}$ is a block diagonal matrix with one $\boldsymbol{\Omega}_{\text{one}}$ for each state dimension.

Using Itô's formula, we can show that the samples of the original Ornstein-Uhlenbeck process $\mathbf{o}$ at each time point can be obtained via the linear coordinate transformation
\begin{equation}
	\mathbf{o}(t) = \mathbf{G}\hat{\mathbf{o}}(t).
\end{equation}
Let $\mathbf{B}$ be defined as the matrix that performs this linear transformation for the unrolled vectors $\mathbf{o} = \mathbf{B} \hat{\mathbf{o}}$.
We can then write the density of the original OU process as 
\begin{equation}
	p(\mathbf{o} \mid \mathbf{G}) = \mathcal{N}\left( \mathbf{o} \mid \mathbf{0}, \mathbf{B} \boldsymbol{\Omega} \mathbf{B}^\top \right).
	\label{eq:OUDensity}
\end{equation}

\subsubsection{Marginals of the Observations}
Using equation \eqref{eq:ObsSumUnnormalized}, the marginal distribution of $\mathbf{y}$ can be computed as the sum of three independent Gaussian-distributed random variables with zero mean, described respectively by equations \eqref{eq:GPPrior}, \eqref{eq:ErrorDensity} and \eqref{eq:OUDensity}. Thus, $\mathbf{y}$ is again Gaussian-distributed, according to
\begin{equation}
	p(\tilde{\mathbf{y}} \mid \bm{\phi}, \mathbf{G}, \bm{\sigma}) = \mathcal{N}(\mathbf{y} \mid \mathbf{0}, \bm{\Sigma}),
	\label{eq:obs_marginals}
\end{equation}
where
\begin{equation}
\bm{\Sigma} = \mathbf{C}_{\bm{\phi}} + \mathbf{T} + \mathbf{B} \bm{\Omega} \mathbf{B}^T.
\end{equation}
Thanks to the latent state representation, the diffusion matrix $\mathbf{G}$ is now a part of the hyperparameters of the observation model. It can then be inferred alongside the hyperparameters of the GP using maximum evidence \citep{rasmussen2004gaussian}. Using a stationary kernel $k$, $\mathbf{C}_{\bm{\phi}} + \mathbf{T}$ captures the stationary part of $\mathbf{z}$ as in standard GP regression, while the parameters in $\mathbf{G}$ describe the non-stationary part due deriving from $\bm{\Omega}$. This ultimately leads to an identifiable problem.

\begin{figure}
	\centering
	\scalebox{.8}{
		\begin{tabular}[c]{cc}
			\begin{subfigure} [t]{.25\textwidth}
				\centering
				\begin{tikzpicture}[
				node distance=0.5 cm and 0.5 cm,
				mynode/.style={draw,circle,minimum size=0.9cm}  % needs to be that big for F_1 to fit
				]%[
				%node distance=0.4 cm and 0.4 cm,
				%]
				\node[mynode](G){$\mathbf{G}$};
				\node[mynode, right= of G](o){$\mathbf{o}$ };
				\node[mynode, right= of o](sigma){$\boldsymbol{\sigma}$};
				\node[mynode, below= of G](zDot){$\mathbf{\dot{z}}$};
				\node[mynode, below= of o](z){$\mathbf{z}$};
				\node[mynode, below= of sigma](y){$\mathbf{y}$};
				\node[mynode, below= of z] (phi) {$\boldsymbol{\phi}$};
				\node[mynode, below= of zDot](theta){$\boldsymbol{\theta}$};
				\path
				(G) edge[-latex] (o)
				(sigma) edge[-latex] (y)
				(phi) edge[-latex] (z)
				(o) edge[-latex] (y)
				(o) edge[-latex] (zDot)
				(z) edge[-latex] (zDot)
				(theta) edge[-latex] (zDot)
				(z) edge[-latex] (y)
				;
				\end{tikzpicture}
				\caption{SDE-based model}
				\label{subfig:SDEModel}
			\end{subfigure} &
			\begin{subfigure} [t]{.25\textwidth}
				\centering
				\begin{tikzpicture}[
				node distance=0.5 cm and 0.5 cm,
				mynode/.style={draw,circle,minimum size=0.9cm}  % needs to be that big for F_1 to fit
				]%[
				%node distance=0.4 cm and 0.4 cm,
				%]
				\node[mynode](G){$\mathbf{G}$};
				\node[mynode, right= of G](o){$\mathbf{o}$ };
				\node[mynode, right= of o](sigma){$\boldsymbol{\sigma}$};
				\node[mynode, below= of G](zDot){$\mathbf{\dot{z}}$};
				\node[mynode, below= of o](z){$\mathbf{z}$};
				\node[mynode, below= of sigma](y){$\mathbf{y}$};
				\node[mynode, below= of z] (phi) {$\boldsymbol{\phi}$};
				\path
				(G) edge[-latex] (o)
				(sigma) edge[-latex] (y)
				(phi) edge[-latex] (z)
				(o) edge[-latex] (y)
				(z) edge[-latex] (zDot)
				(z) edge[-latex] (y)
				(phi) edge[-latex] (zDot)
				;
				\end{tikzpicture}
				\caption{Data-based model}
				\label{subfig:DataModel}
			\end{subfigure}
		\end{tabular}
	}
	\caption{Generative models for the two different ways to compute the derivatives of the latent states $\mathbf{z}$.}
	\label{fig:GenerativeModels}
\end{figure}
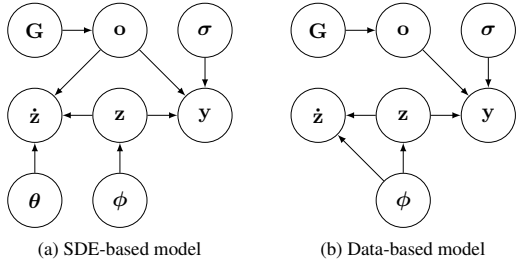

\subsection{Generative Model for Derivatives}

Similarly to gradient matching approaches, we define two generative models for the derivatives $\dot{\mathbf{z}}$, one based on the data and one based on the SDE model.

\subsubsection{Data-based Model}
As shown e.g. in the appendix of \citet{wenk2018fast}, the prior defined in equation \eqref{eq:GPPrior} automatically induces a GP prior on the conditional derivatives of $\mathbf{z}$. Defining
\begin{align}
	\mathbf{D} &\coloneqq {'\mathbf{C}_{\bm{\phi}}} \mathbf{C}_{\bm{\phi}}^{-1},\\
	\mathbf{A} &\coloneqq \mathbf{C}_{\bm{\phi}}'' - {'\mathbf{C}_{\bm{\phi}}} \mathbf{C}_{\bm{\phi}}^{-1} \mathbf{C}_{\bm{\phi}}'
\end{align}
where
\begin{align}
	\left['\mathbf{C}_{\bm{\phi}}\right]_{i,j} &\coloneqq \frac{\partial}{\partial a} k_{\bm{\phi}}(a, b) \bigg\rvert_{a=t_i, b=t_j}, \\
	\left[\mathbf{C}_{\bm{\phi}}'\right]_{i,j} &\coloneqq \frac{\partial}{\partial b} k_{\bm{\phi}}(a, b) \bigg\rvert_{a=t_i, b=t_j}, \\
	\left[\mathbf{C}_{\bm{\phi}}''\right]_{i,j} &\coloneqq \frac{\partial^2}{\partial a \partial b} k_{\bm{\phi}}(a, b) \bigg\rvert_{a=t_i, b=t_j},
\end{align}
we can write
\begin{equation}
	p(\dot{{\mathbf{z}}} \mid {\mathbf{z}}, \bm{\phi}) = \mathcal{N} \left(\dot{{\mathbf{z}}} \mid \mathbf{D} {\mathbf{z}}, \mathbf{A} \right).
\end{equation}

\subsubsection{SDE-based Model}
There also is a second way of obtaining an expression for the derivatives of $\mathbf{{z}}$, namely using equation \eqref{eq:latentDerivs}:
\begin{equation}
	p(\dot{{\mathbf{z}}} \mid \mathbf{o}, {\mathbf{z}}, \bm{\theta}) = \delta(\dot{{\mathbf{z}}} - \mathbf{f}(\mathbf{z} + \mathbf{o}, \bm{\theta}) - \mathbf{o}),
\end{equation}
where $\delta$ represents the dirac delta.

\subsection{Inference}
Combined with the modeling paradigms introduced in the previous sections, this yields the two generative models for the observations in Figure \ref{fig:GenerativeModels}. The graphical model in Figure \ref{subfig:SDEModel} represents the derivatives we get via the generative process described by the SDEs, in particular the nonlinear drift function $\mathbf{f}$. The model in Figure \ref{subfig:DataModel} represents the derivatives yielded by the generative process described by the Gaussian process. Assuming a perfect GP fit and access to the true parameters $\bm{\theta}$, intuitively these two distributions should be equal. We thus want to find parameters $\bm{\theta}$ that minimizes the difference between these two distributions.

Compared to the deterministic ODE case, the graphical models in Figure \ref{fig:GenerativeModels} contain additional dependencies on the contribution of the OU process $\mathbf{o}$. Furthermore, the SDE-driven probability distribution of $\dot{\mathbf{z}}$ in Figure \ref{subfig:SDEModel} depends on $\mathbf{z}$, $\mathbf{o}$, and $\boldsymbol{\theta}$ through a potentially highly nonlinear drift function $\mathbf{f}$. Thus, one cannot do analytical inference without making restrictive assumptions on the functional form of $\mathbf{f}$.

However, as shown in section \ref{sec:AncestralSamplingDensitiesAppendix} of the appendix , it is possible to derive computationally efficient ancestral sampling schemes for both models, as summarized in Algorithm \ref{samplingAlgo}. While this rules out classical approaches like analytically minimizing the KL divergence, we can now deploy likelihood-free algorithms that were designed for matching two probability densities based on samples.
\begin{algorithm}
	\caption{Ancestral sampling for $\mathbf{\dot{z}}$}
	\begin{algorithmic}[1]
		\State{\textbf{Input:}\quad  $\mathbf{y}, \mathbf{f}(\mathbf{z}, \bm{\theta}), \mathbf{t}, \bm{\sigma}, \mathbf{G}$}
		\State \emph{Ancestral sampling the SDE model}
		\State $\textrm{Sample } \mathbf{o}_\text{s} \textrm{ by drawing from } p(\mathbf{o} \mid \mathbf{G})$
		\State $\textrm{Sample } \mathbf{z}_\text{s} \textrm{ by drawing from } p(\mathbf{z} \mid \mathbf{y}, \mathbf{o}, \bm{\sigma}) \textrm{ using } \mathbf{o}_\text{s}$
		\State $\textrm{Sample } \dot{\mathbf{z}}_\text{s} \textrm{ by drawing from } p(\dot{\mathbf{z}} \mid \mathbf{o}, \mathbf{z}, \bm{\theta}) \textrm{ using } \mathbf{o}_\text{s}, \mathbf{z}_\text{s}$
		\State \emph{Ancestral sampling the Data model}
		\State $\textrm{Sample } \mathbf{z}_\text{d} \textrm{ by drawing from } p(\mathbf{z} \mid \mathbf{y}, \mathbf{G}, \bm{\sigma})$
		\State $\textrm{Sample } \dot{\mathbf{z}}_\text{d} \textrm{ by drawing from } p(\dot{\mathbf{z}} \mid \mathbf{z}, \bm{\phi})$
		\State{\textbf{Return:} $\dot{\mathbf{z}}_\text{s}, \dot{\mathbf{z}}_\text{d}$}
	\end{algorithmic}
	\label{samplingAlgo}
\end{algorithm}

\subsection{Adversarial Sample-based Inference}
\begin{figure}
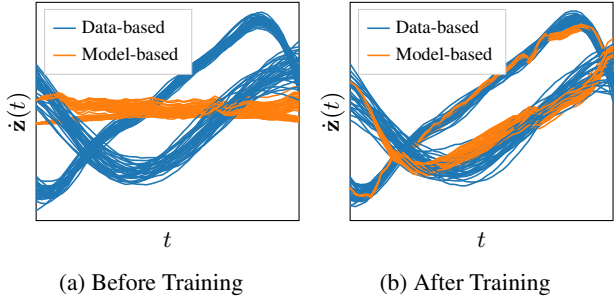

	\begin{subfigure}[c]{0.238\textwidth}
		\centering
		\input{plots/gradients_lv_training/lotka_volterra_before.tex}
		\vspace{-1\baselineskip}
		\caption{Before Training}
	\end{subfigure}
	\hfill
	\begin{subfigure}[c]{0.238\textwidth}
		\centering
		\input{plots/gradients_lv_training/lotka_volterra_after.tex}
		\vspace{-1\baselineskip}
		\caption{After Training}
	\end{subfigure}
	\caption{Comparing gradients sampled from the graphical model in Figure \ref{subfig:SDEModel} (Model-based) and the graphical model in Figure \ref{subfig:DataModel} (Data-based) before and after adversarial training on Lotka Volterra.}
	\label{fig:samples}
\end{figure}
Arguably, generative adversarial networks (GANs) \citep{goodfellow2014generative} are amongst the most popular algorithms of this kind; here a parametric neural network is trained to match the unknown likelihood of the data. The basic GAN setup consists of a fixed data set, a generator that tries to create realistic samples of said dataset and a discriminator that tries to tell apart the fake samples from the true ones. As recently shown by \citet{yang2018physics}, GANs have the potential to solve stochastic partial differential equations (SPDEs). \citet{yang2018physics} assume a fixed data set consisting of observations (similar to the $\mathbf{y}$ in this paper) and use an SPDE-inspired neural network as a generator for realistic observations. In the case of SDEs however, this would still involve a lot of numerical integration. Thus, we modify the GAN setup by leaving behind the idea of having a fixed data set. Instead of relying on bootstrapped samples of the observations $\mathbf{y}$, we sample the derivatives from the data-based model shown in Figure \ref{subfig:DataModel}. For a sufficiently good model fit, these samples represent the true derivatives of the latent variable $\mathbf{z}$. We then use the SDE-based model shown in Figure \ref{subfig:SDEModel} as a generator. To avoid standard GAN problems such as training instability and to improve robustness, we choose to replace the discriminator with a critic $\mathcal{C}_\omega$. As proposed by \citet{arjovsky2017wasserstein}, this critic is trained to estimate the Wasserstein distance between the derivative samples. The resulting algorithm, summarized in Algorithm \ref{algo:wgan_algorithm}, can be interpreted as performing Adversarial Regression for SDEs and will thus be called AReS. In Figure \ref{fig:samples}, we show the derivatives sampled from the two models both before and after training for one example run of the Lotka Volterra system (cf. Section \ref{subsec:nonDiagDiff}). While not perfect, the GAN is clearly able to push the SDE gradients towards the gradients of the observed data.

\subsection{Maximum Mean Discrepancy}

Even though they work well in practical settings, during training GANs need ad hoc precautions and careful balancing between their generator and discriminator. \citet{dziugaite2015training} propose to solve this problem using Maximum Mean Discrepancy (MMD) \citep{gretton2012kernel} as a metric to substitute the discriminator. As proposed by \citet{li2015generative}, we choose the rational quadratic kernel to obtain a robust discriminator that can be deployed without fine-tuning on a variety of problems. The resulting procedure, summarized in Algorithm \ref{algo:mmd_algorithm}, can be interpreted as performing Maximum mean discrepancy-minimizing Regression for SDEs and will thus be called MaRS.
\begin{algorithm}[!h]
	\caption{AReS}
	\begin{algorithmic}[1]
		\State{\textbf{Input:}\quad  Observations $\mathbf{y}$ at times $\mathbf{t}$, a model $\mathbf{f}$, learning rate $\alpha$, number of total iterations $N_{\text{it}}$, the clipping parameter $c$, the batch size $M$, the number of iterations of the critic per generator iteration $n_{\text{critic}}$.
			\State Train the Gaussian process on the data to recover the hyperparameters $\bm{\phi}$, $\bm{\sigma}$ and the diffusion $\mathbf{G}$
			\State Initialize the critic parameters $\bm{\omega}$ and the SDE parameters $\bm{\theta}$ respectively with $\bm{\omega}_0$ and $\bm{\theta}_0$}
		\For{$n_\text{it} = 1, \dots, N_{\text{it}} $ }
		\For{$n_\text{c} = 1, \dots, n_{\text{critic}}$}
		\State Sample $\dot{\mathbf{z}}_\text{s} \sim p_{\text{s}}(\dot{\mathbf{z}})$ and $\dot{\mathbf{z}}_\text{d} \sim p_{\text{d}}(\dot{\mathbf{z}})$ as described in algorithm \ref{samplingAlgo}. Each batch contains $M$ elements
		\State $g_{\bm{\omega}} \leftarrow { \scriptstyle \nabla_{\bm{\omega}} \left[ \frac{1}{M} \sum_{i=1}^{M} \mathcal{C}_{\bm{\omega}}(\dot{\mathbf{z}}_\text{d}^{(i)}) - \frac{1}{M} \sum_{i=1}^{M} \mathcal{C}_{\bm{\omega}}(\dot{\mathbf{z}}_\text{s}^{(i)}) \right]}$
		\State $\bm{\omega} \leftarrow \bm{\omega} + \alpha \cdot \text{Adam}(\bm{\omega}, g_{\bm{\omega}})$
		\State $\bm{\omega} \leftarrow \text{clip}(\bm{\omega}, -c, c)$
		\EndFor
		\State $g_{\bm{\theta}} \leftarrow {\scriptstyle - \nabla_{\bm{\theta}} \frac{1}{M} \sum_{i=1}^{M} f_{\bm{\omega}}(\dot{\mathbf{z}}_\text{s}^{(i)})}$
		\State $\bm{\theta} \leftarrow \bm{\theta} - \alpha \cdot \text{Adam}(\bm{\theta}, g_{\bm{\theta}})$
		\EndFor
	\end{algorithmic}
	\label{algo:wgan_algorithm}
\end{algorithm}
\begin{algorithm}[!h]
	\caption{MaRS}
	\begin{algorithmic}[1]
		\State{\textbf{Input:}\quad  Observations $\mathbf{y}$ at times $\mathbf{t}$, SDE model $\mathbf{f}$, learning rate $\alpha$, number of iterations $N_{\text{it}}$, batch size $M$}
		\State Train the Gaussian process on the data to recover the hyperparameters $\bm{\phi}$, $\bm{\sigma}$ and the diffusion $\mathbf{G}$
		\State Initialize the SDE parameters with $\bm{\theta}_0$ 
		\For{$n_\text{it} = 1, \dots, N_{\text{it}} $ }
		\State Sample $\dot{\mathbf{z}}_\text{s} \sim p_{\text{s}}(\dot{\mathbf{z}})$ and $\dot{\mathbf{z}}_\text{d} \sim p_{\text{d}}(\dot{\mathbf{z}})$ as described in algorithm \ref{samplingAlgo}. Each batch contains $M$ elements
		\State $g_{\bm{\theta}} \leftarrow \nabla_{\bm{\theta}} MMD^2_u \left[\dot{\mathbf{z}}_\text{s}, \dot{\mathbf{z}}_\text{d} \right]$
		\State $\bm{\theta} \leftarrow \bm{\theta} - \alpha \cdot \text{Adam}( \bm{\theta}, g_{\bm{\theta}})$
		\EndFor
	\end{algorithmic}
	\label{algo:mmd_algorithm}
\end{algorithm}

\section{Experiments}
\label{sec:experiments}
% experiments

\subsection{Setups}
\label{sec:setups}
To evaluate the empirical performance of our method, we conduct several experiments on simulated data, using four standard benchmark systems and comparing against the EKF-based approach by \citet{sarkka2015posteriorCOMPARISON} and two GP-based approaches respectively by \citet{vrettas2015variational} and \citet{yildiz2018learning}.

\begin{figure}
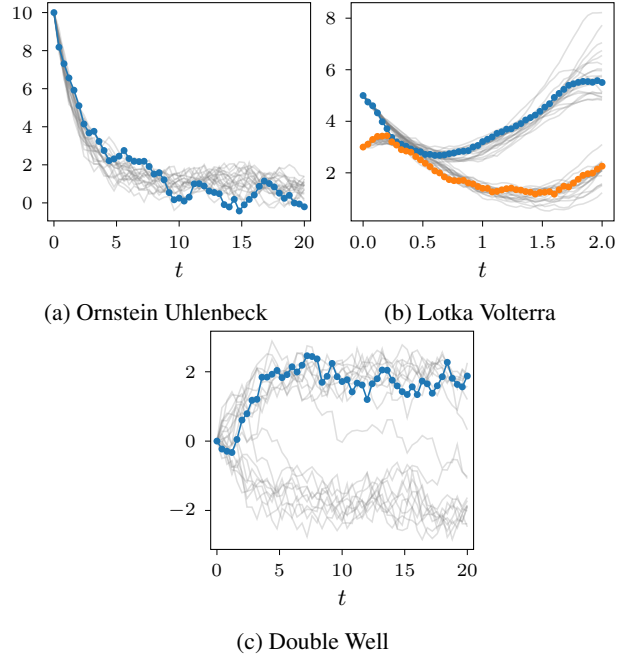

	\begin{center}
		\begin{subfigure}[t]{0.238\textwidth}
			\centering
			\input{plots/sample_trajectories/ornstein_uhlenbeck_samples.tex}
			\vspace{-1\baselineskip}
			\caption{Ornstein Uhlenbeck}
			\label{subfig:OU_Dynamics}
		\end{subfigure}
		\hfill
		\begin{subfigure}[t]{0.238\textwidth}
			\centering
			\input{plots/sample_trajectories/lotka_volterra_samples.tex}
			\vspace{-1\baselineskip}
			\caption{Lotka Volterra}
			\label{subfig:LV_Dynamics}
		\end{subfigure}
		\hfill
		\begin{subfigure}[c]{0.238\textwidth}
			\centering
			\input{plots/sample_trajectories/double_well_samples.tex}
			\vspace{-1\baselineskip}
			\caption{Double Well}
			\label{subfig:DW_Dynamics}
		\end{subfigure}
	\end{center}
	\caption{Sample trajectories for three different benchmark systems. While Ornstein Uhlenbeck and Lotka Volterra are rather tame, the Double Well potential clearly exhibits a bifurcation effect.}
\end{figure}

The first system is a simple Ornstein-Uhlenbeck process as shown in Figure \ref{subfig:OU_Dynamics}, given by the SDE
\begin{equation}
	\mathrm{d} x(t) = \theta_0 ( \theta_1 - x(t)) \mathrm{d}t + G \mathrm{d}w(t).
\end{equation}
As mentioned in Section \ref{sec:OUContributions}, this system has an analytical Gaussian process solution and thus serves more academic purposes. We use $\bm{\theta} = [0.5, 1.0]$, $G = 0.5$ and $x(0) = 10$.

The second system is the Lorenz '63 model given by the SDEs
\begin{align*}
	\mathrm{d} x_1(t) &= \theta_1 (x_2(t) - x_1(t))\mathrm{d} t && + \sigma_1 \mathrm{d} w_1(t)\\
	\mathrm{d} x_2(t) &= (\theta_2 x_1(t) - x_2(t) - x_1(t)x_3(t)) \mathrm{d} t && + \sigma_2 \mathrm{d} w_2(t)\\
	\mathrm{d} x_3(t) &= (x_1(t)x_2(t) - \theta_3 x_3(t))\mathrm{d} t && + \sigma_3 \mathrm{d} w_3(t).
\end{align*}
In both systems, the drift function is linear in one state or one parameter conditioned on all the others \citep[cf.][]{gorbach2017scalable}. Furthermore, there is no coupling across state dimensions in the diffusion matrix. This leads to two more interesting test cases.

To investigate the algorithm's capability to deal with off-diagonal terms in the diffusion, we introduce the two dimensional Lotka-Volterra system shown in Figure \ref{subfig:LV_Dynamics}, given by the SDEs
\begin{equation}
\mathrm{d}\mathbf{x}(t) = \left[ \begin{matrix}
\theta_1 x_1(t) - \theta_2 x_1(t) x_2(t)\\
-\theta_3 x_2(t) + \theta_4 x_1(t) x_2(t) 
\end{matrix} \right] \mathrm{d}t + \mathbf{G}\mathrm{d}\mathbf{w}(t),
\end{equation}
where $\mathbf{G}$ is, without loss of generality, assumed to be a lower triangular matrix. The true vector parameter is $\bm{\theta} = [2,1,4,1]$ and the system is simulated starting from $ \mathbf{x}(0) = [3, 5]$. Since its original introduction by \citet{lotka1932growth}, the Lotka Volterra system has been widely used to model population dynamics in biology. The system is observed at 50 equidistant points in the interval $t = [0,\ 20]$. As it turns out, this problem is significantly challenging for all algorithms, despite the absence of observation noise.

To investigate the effect of strong non-linearities in the drift, we introduce the Ginzburg-Landau double-well potential shown in Figure \ref{subfig:DW_Dynamics}, defined by the SDE
\begin{equation}
	\mathrm{d}x(t) = \theta_0 x ( \theta_1 - x^2)\mathrm{d}t + G \mathrm{d}w(t).
\end{equation}
Using $\bm{\theta} = [0.1, 4]$, $G = 0.5$ and $x(0)=0$, this system exhibits an interesting bifurcation effect. While there are two stable equilibria at $x=\pm2$, the one the system will end up in is completely up to noise. For this reason it represents a fitting framework to test how well an algorithm can deal with multi-modal SDEs.
The potential value is observed at 50 equidistant points in the interval $t = [0,\ 20]$, subjected to observational noise with $\sigma = 0.2$.

Lastly, some implementation details are constant throughout each experiment: the critic in the adversarial parameter estimation is a 2-layer fully connected neural network, with respectively 256 and 128 nodes. Every batch, for both MMD and adversarial training contains 256 elements. While the Ornstein-Uhlenbeck process and the double-well potential were modeled with a sigmoid kernel, for Lotka-Volterra and Lorenz '63 we used a common RBF (we point at \citet{rasmussen2004gaussian} for more information about kernels and GPs).

\subsection{Evaluation}
For all systems, the parameters $\bm{\theta}$ turn out to be identifiable. Thus, the parameter value is a good indicator of how well an algorithm is able to infer the drift function. However, since the components of $\mathrm{d}\mathbf{w}(t)$ are independent, there are multiple diffusion matrices $\mathbf{G}$ that generate the same trajectories. We thus directly compare the variance of the increments, i.e. the elements of $\mathbf{H} \coloneqq \mathbf{G}^{T} \mathbf{G}$.

To account for statistical fluctuations, we use 100 independent realizations of the SDE systems and compare the mean and standard deviation of $\bm{\theta}$ and $\mathbf{H}$. $\mathrm{AReS}$ and $\mathrm{MaRS}$ are compared against the Gaussian process-based $\mathrm{VGPA}$ by \citet{vrettas2015variational} and $\mathrm{NPSDE}$ by \citet{yildiz2018learning} as well as the classic Kalman filter-based $\mathrm{ESGF}$ recommended by \citet{sarkka2015posteriorCOMPARISON}.

\subsection{Locally Linear Systems}
As mentioned in Section \ref{sec:setups}, the functional form of the drift functions of both the Ornstein-Uhlenbeck process and the Lorenz '63 system satisfies a local linearity assumption, while their diffusion is kept diagonal. Thus, they serve as excellent benchmarks for parameter inference algorithms. The empirical results are shown in Table \ref{tab:OU} for the Ornstein-Uhlenbeck. Unfortunately, $\mathrm{VGPA}$ turns out to be rather unstable if both diffusion and parameters are unknown, despite on average roughly 54 hours are needed to observe convergence. We then provide it with the true $\mathbf{G}$ and show only its empirical parameter estimates. Since both $\mathrm{AReS}$ and $\mathrm{MaRS}$ use Equation \eqref{eq:obs_marginals} to determine $\mathbf{G}$, they share the same values. Due to space restrictions, the results for Lorenz '63 can be found in Table \ref{tab:Lorenz63} of the appendix. As demonstrated by this experiment, $\mathrm{AReS}$ and $\mathrm{MaRS}$ can deal with locally linear systems, outperforming their competitors, especially in their estimates of the diffusion terms.

\subsection{Non-Diagonal Diffusion}
\label{subsec:nonDiagDiff}
To investigate the effect of off-diagonal entries in $\mathbf{G}$, we use the Lotka-Volterra dynamics. Since $\mathrm{NPSDE}$ is unable to model non-diagonal diffusions, we provide it with the true $\mathbf{G}$ and only compare parameter estimates. As $\mathrm{VGPA}$ is already struggling in the lower dimensional cases, we omit it from this comparison due to limited computational resources. The results are shown in Table \ref{tab:LV}. $\mathrm{AReS}$ and $\mathrm{MaRS}$ clearly outperform the other methods in terms of diffusion estimation, while $\mathrm{ESGF}$ is the only algorithm that yields drift parameter estimates of comparable quality.

\subsection{Dealing with Multi-Modality}
\begin{table*}
	\caption{Inferred parameters over 100 independent realizations of respectively the Ornstein-Uhlenbeck, Ginzburg-Landau Double-Well and Lotka-Volterra dynamics. For every algorithm, we show the median $\pm$ one standard deviation.\vspace{0.1cm}}
	\begin{subtable}{\textwidth}
		\begin{center}
			\scalebox{.9}{
			\begin{tabular}{ |l|c|c|c|c|c| }
					\hline
					Ground truth & $\mathrm{NPSDE}$ & $\mathrm{VGPA}$ & $\mathrm{ESGF}$ & $\mathrm{AReS}$ & $\mathrm{MaRS}$ \\
					\hline
					$\theta_0 = 0.5$ & $0.41 \pm 0.11$ & $0.53 \pm 0.08$ & $0.49 \pm 0.07$ & $\mathbf{0.50 \pm 0.21}$ & $0.46 \pm 0.06$ \\
					\hline
					$\theta_1 = 1$ & $0.71\pm 1.34$ & $0.96 \pm 0.31$ & $0.96 \pm 0.24$ & $1.06 \pm 0.93$ & $\mathbf{0.99 \pm 0.25}$ \\ 
					\hline
					$H = 0.25$ & $0.00 \pm 0.01$ & / & $0.19 \pm 0.06$ & \multicolumn{2}{|c|}{$\mathbf{0.24 \pm 0.09}$} \\
					\hline
			\end{tabular} }
		\end{center}
		\caption{Ornstein-Uhlenbeck process}
		\label{tab:OU}
	\end{subtable}
	% ------------------------------------------------------------
	\begin{subtable}{\textwidth}
		\begin{center}
			\scalebox{.9}{
				\begin{tabular}{ |l|c|c|c|c | }
					\hline
					Ground truth & $\mathrm{NPSDE}$ & $\mathrm{ESGF}$ & $\mathrm{AReS}$ & $\mathrm{MaRS}$ \\
					\hline
					$\theta_0 = 2$ & $1.58 \pm 0.71$ & $2.04 \pm 0.09$ & $2.36 \pm 0.18$ & $\mathbf{2.00 \pm 0.09}$ \\
					\hline
					$\theta_1 = 1$ & $0.74 \pm 0.31$ & $1.02 \pm 0.05$ & $1.18 \pm 0.9$ & $\mathbf{1.00 \pm 0.04}$ \\
					\hline
					$\theta_2 = 4$ & $2.26 \pm 1.51$ & $3.87 \pm 0.59$ & $\mathbf{3.97 \pm 0.63}$ & $3.70 \pm 0.51$ \\ 
					\hline
					$\theta_3 = 1$ & $0.49 \pm 0.35$ & $0.96 \pm 0.14$ & $\mathbf{0.98 \pm 0.18}$ & $0.91 \pm 0.14$ \\ 
					\hline
					$\mathbf{H}_{1,1} = 0.05$ & / & $0.01 \pm 0.03$ & \multicolumn{2}{|c|}{$\mathbf{0.03 \pm 0.004}$} \\
					\hline
					$\mathbf{H}_{1,2} = 0.03$ & / & $0.01 \pm 0.01$ & \multicolumn{2}{|c|}{$\mathbf{0.02 \pm 0.01}$} \\
					\hline
					$\mathbf{H}_{2,1} = 0.03$ & / & $0.01 \pm 0.01$ & \multicolumn{2}{|c|}{$\mathbf{0.02 \pm 0.01}$} \\ 
					\hline
					$\mathbf{H}_{2,2} = 0.09$ & / & $0.03 \pm 0.02$ & \multicolumn{2}{|c|}{$\mathbf{0.09 \pm 0.03}$} \\ 
					\hline
				\end{tabular}}
		\end{center}
		\caption{Lotka-Volterra}
		\label{tab:LV}
	\end{subtable}
	% ------------------------------------------------------------
	\begin{subtable}{\textwidth}
		\begin{center}
			\scalebox{.9}{
				\begin{tabular}{ |l|c|c|c|c|c| }
					\hline
					Ground truth & $\mathrm{NPSDE}$ & $\mathrm{VGPA}$ & $\mathrm{ESGF}$ & $\mathrm{AReS}$ & $\mathrm{MaRS}$ \\
					\hline
					$\theta_0 = 0.1$ & $0.09 \pm 7.00 $ & $0.05 \pm 0.04$ & $0.01 \pm 0.03$ & $0.09 \pm 0.04$ & $\mathbf{0.10 \pm 0.05}$ \\
					\hline
					$\theta_1 = 4$ & $3.36\pm 248.82$ & $1.11 \pm 0.66$ & $0.11 \pm 0.16$ & $3.68 \pm 1.34$ & $\mathbf{3.85 \pm 1.10}$\\ 
					\hline
					$H = 0.25$ & $0.00 \pm 0.02$& / & $0.20 \pm 0.05$ & \multicolumn{2}{|c|}{$\mathbf{0.21 \pm 0.09}$} \\
					\hline
			\end{tabular} }
		\end{center}
		\caption{Double-Well potential}
		\label{tab:DW}
	\end{subtable}
\end{table*}
As a final challenge, we investigate the Ginzburg-Landau double well potential. Despite one-dimensional, its state distribution is multi-modal even if all parameters are known. As shown in Table \ref{tab:DW}, this is definitely a challenge for all classical approaches. While the number of data-points is probably not enough for the non-parametric proxy for the drift function in $\mathrm{NPSDE}$, the time-dependent Gaussianity assumptions in both $\mathrm{VGPA}$ and $\mathrm{ESGF}$ are problematic in this case. In our gradient matching framework, no such assumption is made. Thus, both $\mathrm{AReS}$ and $\mathrm{MaRS}$ are able to deal with the multimodality of the problem.

\section{Conclusion}
Parameter and diffusion estimations in stochastic systems arise in quantitative sciences and many fields of engineering. Current techniques based on Kalman filtering or Gaussian processes approximate the state distribution conditioned on the parameters and iteratively optimize the data likelihood. In this work, we propose to turn this procedure on its head by leveraging key ideas from gradient matching algorithms, originally designed for deterministic ODEs. By introducing a novel noise model for Gaussian process regression that leverages the Doss-Sussmann transformation, we are able to reliably estimate the parameters in the drift and the diffusion processes. Our algorithm can keep up with and occasionally outperform the state-of-the-art on the simpler benchmark systems, while it is also accurately estimating parameters for systems that exhibit multi-modal state densities, a case where traditional methods fail. While our approach is currently restricted to systems with a constant diffusion matrix $\mathbf{G}$, it would be interesting to see how it generalizes to other settings, perhaps using alternative or approximate bridge constructs. Unfortunately, this is outside of the scope of this work. We hope nevertheless that the publicly available code will facilitate future research in that direction.

\section*{Acknowledgements}
This research was supported by the Max Planck ETH Center for Learning Systems. GA acknowledges funding from Google DeepMind and University of Oxford.
This project has received funding from the European Research Council (ERC) under the European Union’s Horizon 2020 research and innovation programme grant agreement No 815943.
%If a paper is accepted, the final camera-ready version can (and
%probably should) include acknowledgements. In this case, please
%place such acknowledgements in an unnumbered section at the
%end of the paper. Typically, this will include thanks to reviewers
%who gave useful comments, to colleagues who contributed to the ideas,
%and to funding agencies and corporate sponsors that provided financial
%support.

% Bibliography
\bibliography{references}
\bibliographystyle{icml2019}

\appendix
\onecolumn
\section{Supplementary Material}
\label{sec:supplement}
\subsection{Parameter Estimation Lorenz '63}

\begin{table*}[h]
	\centering
	\begin{tabular}{ |l|c|c|c|c|}
		\hline
		Ground truth & $\mathrm{NPSDE}$ & $\mathrm{ESGF}$ & $\mathrm{AReS}$ & $\mathrm{MaRS}$ \\
		\hline
		$\theta_0 = 10$  & $1.28 \pm 2.32$ & $\mathbf{9.97 \pm 0.33}$ & $7.24 \pm 1.08$ & $9.82 \pm 0.56$ \\
		\hline
		$\theta_1 = 28$  & $20.69 \pm 5.73$ & $\mathbf{28.00 \pm 0.17}$ & $28.16 \pm 1.08$ & $27.96 \pm 0.21$ \\
		\hline
		$\theta_0 = 2.667$  & $1.86 \pm 1.08$ & $\mathbf{2.65 \pm 0.06}$ & $2.55 \pm 0.10$ & $2.64 \pm 0.07$ \\	
		\hline
		$G = \sqrt{10}$ & $6.51 \pm 1.31$ & $\mathbf{3.03 \pm 0.2}$ &  \multicolumn{2}{|c|}{$3.54 \pm 2.45$} \\	
		\hline	
	\end{tabular}
	\caption{Median and standard deviation of the 65 best runs of each algorithm. As $\mathrm{ESGF}$ crashed in roughly one third of all experiments, we compare only the best 65 runs, where a crash is treated as a complete failure. While this provides somehow a fair comparison, it should be noted that this significantly overestimates the performance of all algorithms.}
	\label{tab:Lorenz63}
\end{table*}

\subsection{Training Times}

\begin{table*}[h]
	\centering
	\begin{tabular}{ |l|c|c|c|c|c|}
		\hline
		& $\mathrm{NPSDE}$ & $\mathrm{VGPA}$ & $\mathrm{ESGF}$ & $\mathrm{AReS}$ & $\mathrm{MaRS}$ \\
		\hline
		OU Process  & $48.8 \pm 0.9$ & $\sim(54 \pm 8)$hours & $32.2 \pm 0.3$ & $321.3 \pm 0.8$ & $17.3 \pm 0.3$ \\
		\hline
		DW Potential  & $406.9 \pm 147.9$ & $\sim(12 \pm 6)$hours  & $35.2 \pm 0.1$ & $326.0 \pm 1.8$ & $17.7 \pm 2.0$\\
		\hline
		Lotka-Volterra  & $1421.8 \pm 1.0$ & / & $244.7 \pm 1.2$ & $47.6 \pm 1.3$ & $19.3 \pm 1.1$ \\	
		\hline
		Lorenz '63 & $39273.5 \pm 8.9$ & / & $670.7 \pm 10.7$ & $26274.0 \pm 2529.8$ & $721.1 \pm 10.7$ \\	
		\hline	
	\end{tabular}
	\caption{Computational times (in seconds) required for training the different algorithms.}
	\label{tab:training_times}
\end{table*}

\subsection{Densities for Ancestral Sampling of the SDE-Based Model}
\label{sec:AncestralSamplingDensitiesAppendix}

Given the graphical model in Figure \ref{subfig:SDEModel}, it is straightforward to compute the densities used in the ancestral sampling scheme in Algorithm \ref{samplingAlgo}. After marginalizing out $\mathbf{\dot{z}}$, the joint density described by the graphical model can be written as
\begin{align}
	p(\mathbf{o}, \mathbf{z}, \mathbf{y} | \boldsymbol{\phi}, \mathbf{G}, \boldsymbol{\sigma}) &= p(\mathbf{o} | \mathbf{G})p(\mathbf{z} | \boldsymbol{\phi})p(\mathbf{y} | \mathbf{z}, \mathbf{o}, \boldsymbol{\sigma})
\end{align}
Substituting the densities given by Equations \eqref{eq:ObsSumUnnormalized}, \eqref{eq:GPPrior}, \eqref{eq:ErrorDensity} and \eqref{eq:OUDensity} yields
\begin{align}
	p(\mathbf{o}, \mathbf{z}, \mathbf{y} | \boldsymbol{\phi}, \mathbf{G}, \boldsymbol{\sigma}) &=
	\mathcal{N}(\mathbf{o} | \mathbf{0} ,  \mathbf{B} \boldsymbol{\Omega} \mathbf{B}^T )
	\mathcal{N}(\mathbf{z} | \mathbf{0}, \mathbf{C}_{\boldsymbol{\phi}})
	\mathcal{N}(\mathbf{y} | \mathbf{z} + \mathbf{o} , \mathbf{T}).
\end{align}

Using a change of variables to simplify notation, we write
\begin{align}
	p(\mathbf{o}, \mathbf{z}, \mathbf{y} | \boldsymbol{\phi}, \mathbf{G}, \boldsymbol{\sigma}) &=
	\mathcal{N}(\mathbf{o} | \mathbf{0} , \tilde{\boldsymbol{\Omega}})
	\mathcal{N}(\mathbf{z} | \mathbf{0}, \mathbf{C}_{\boldsymbol{\phi}})
	\mathcal{N}(\mathbf{y} | \mathbf{z} +\mathbf{o} , \mathbf{T}). \label{eq:varSub}
\end{align}
This equation is now subsequently modified by observing that the product of two Gaussian densities in the same random variable is again a Gaussian density:
\begin{align}
	\nonumber
	p(\mathbf{o}, \mathbf{z}, \mathbf{y} | \boldsymbol{\phi}, \mathbf{G}, \boldsymbol{\sigma}) &=
	\mathcal{N}(\mathbf{o} | \mathbf{0} , \tilde{\boldsymbol{\Omega}})
	\mathcal{N}(\mathbf{z} | \mathbf{0}, \mathbf{C}_{\boldsymbol{\phi}})
	\mathcal{N}(\mathbf{y} | \mathbf{z} +\mathbf{o} , \mathbf{T})\\
	\nonumber
	&= \mathcal{N}(\mathbf{o} | \mathbf{0} , \tilde{\boldsymbol{\Omega}})
	\mathcal{N}(\mathbf{z} | \mathbf{0}, \mathbf{C}_{\boldsymbol{\phi}})
	\mathcal{N}(\mathbf{z} | \mathbf{y} - \mathbf{o} , \mathbf{T})\\
	\nonumber
	&= \mathcal{N}(\mathbf{o} | \mathbf{0} , \tilde{\boldsymbol{\Omega}})
	\mathcal{N}(\mathbf{y} - \mathbf{o} | \mathbf{0}, \mathbf{C}_{\boldsymbol{\phi}} + \mathbf{T})
	\mathcal{N}(\mathbf{z} | \mathbf{m}_z , \mathbf{C}_z)\\
	\nonumber
	&= \mathcal{N}(\mathbf{o} | \mathbf{0} , \tilde{\boldsymbol{\Omega}})
	\mathcal{N}(\mathbf{o} | \mathbf{y}, \mathbf{C}_{\boldsymbol{\phi}} + \mathbf{T})
	\mathcal{N}(\mathbf{z} | \mathbf{m}_z , \mathbf{C}_z)\\
	&= \mathcal{N}(\mathbf{y} | \mathbf{0}, \tilde{\boldsymbol{\Omega}} + \mathbf{C}_{\boldsymbol{\phi}} + \mathbf{T})
	\mathcal{N}(\mathbf{o} | \mathbf{m}_o, \mathbf{C}_o)
	\mathcal{N}(\mathbf{z} | \mathbf{m}_z, \mathbf{C}_z)
	\label{eq:jointDensitySDEBased}
\end{align}
where
\begin{align}
\mathbf{m}_z &= \mathbf{C}_z (\mathbf{T}^{-1}(\mathbf{y} - \mathbf{o}))\\
\mathbf{C}_z &= (\mathbf{C}_{\boldsymbol{\phi}}^{-1} + \mathbf{T}^{-1})^{-1}\\
\mathbf{m}_o &= \mathbf{C}_o(\mathbf{C}_{\boldsymbol{\phi}} + \mathbf{T})^{-1} \mathbf{y}\\
\mathbf{C}_o &= (\tilde{\boldsymbol{\Omega}}^{-1} + (\mathbf{C}_{\boldsymbol{\phi}} + \mathbf{T})^{-1})^{-1}
\end{align}

This formula can be further refined with the Woodbury identity, i.e.
\begin{align}
	\nonumber
	\mathbf{C}_z &= (\mathbf{C}_{\boldsymbol{\phi}}^{-1} + \mathbf{T}^{-1})^{-1}\\
	\nonumber
	&=\mathbf{C}_{\boldsymbol{\phi}} - \mathbf{C}_{\boldsymbol{\phi}}(\mathbf{C}_{\boldsymbol{\phi}} + \mathbf{T})^{-1}\mathbf{C}_{\boldsymbol{\phi}}\\
	&=\mathbf{C}_{\boldsymbol{\phi}}(\mathbf{C}_{\boldsymbol{\phi}} + \mathbf{T})^{-1} \mathbf{T}
\end{align}
which leads to
\begin{equation}
	\mathbf{m}_z =\mathbf{C}_{\boldsymbol{\phi}}(\mathbf{C}_{\boldsymbol{\phi}} + \mathbf{T})^{-1}(\mathbf{y} - \mathbf{o})
\end{equation}
and
\begin{align}
	\nonumber
	\mathbf{C}_o &= (\tilde{\boldsymbol{\Omega}}^{-1} + (\mathbf{C}_{\boldsymbol{\phi}} + \mathbf{T})^{-1})^{-1}\\
	\nonumber
	&= \tilde{\boldsymbol{\Omega}} - \tilde{\boldsymbol{\Omega}} (\tilde{\boldsymbol{\Omega}} + \mathbf{C}_{\boldsymbol{\phi}} + \mathbf{T})^{-1} \tilde{\boldsymbol{\Omega}}\\
	&=\tilde{\boldsymbol{\Omega}} (\tilde{\boldsymbol{\Omega}} + \mathbf{C}_{\boldsymbol{\phi}} + \mathbf{T})^{-1}(\mathbf{C}_{\boldsymbol{\phi}} + \mathbf{T})
\end{align}
which leads to
\begin{equation}
	\mathbf{m}_o = \tilde{\boldsymbol{\Omega}} (\tilde{\boldsymbol{\Omega}} + \mathbf{C}_{\boldsymbol{\phi}} + \mathbf{T})^{-1} \mathbf{y}
\end{equation}

Since we observe $\mathbf{y}$, we are interested in calculating the conditional distribution
\begin{equation}
	p(\mathbf{o}, \mathbf{z} | \mathbf{y}, \boldsymbol{\phi}, \mathbf{G}, \boldsymbol{\sigma}) =  \frac{p(\mathbf{o}, \mathbf{z}, \mathbf{y} | \boldsymbol{\phi}, \mathbf{G}, \boldsymbol{\sigma})}{p(\mathbf{y} | \boldsymbol{\phi}, \mathbf{G}. \boldsymbol{\sigma})}
\end{equation}
Conveniently enough, the marginal density of $\mathbf{y}$ is already factorized out in Equation \eqref{eq:jointDensitySDEBased} (compare Equation \eqref{eq:obs_marginals}). Thus, we have
\begin{equation}
	p(\mathbf{o}, \mathbf{z} | \mathbf{y}, \boldsymbol{\phi}, \mathbf{G}, \boldsymbol{\sigma}) =
	\mathcal{N}(\mathbf{o} | \mathbf{m}_o, \mathbf{C}_o)
	\mathcal{N}(\mathbf{z} | \mathbf{m}_z, \mathbf{C}_z)
\end{equation}
As $\mathcal{N}(\mathbf{o} | \mathbf{m}_o, \mathbf{C}_o)$ is independent of $\mathbf{z}$, we can employ ancestral sampling by first obtaining a sample of $\mathbf{o}$ through $\mathcal{N}(\mathbf{o} | \mathbf{m}_o, \mathbf{C}_o)$, and then utilizing such sample to get $\mathbf{z}$ through $\mathcal{N}(\mathbf{z} | \mathbf{m}_z, \mathbf{C}_z)$.

\subsection{Calculating the GP Posterior for Data-Based Ancestral Sampling}
Given the graphical model in Figure \ref{subfig:DataModel}, we can calculate the densities used in the ancestral sampling scheme in Algorithm \ref{samplingAlgo}. After marginalizing out $\mathbf{\dot{z}}$ and using the variable substitutions introduced in Equation \eqref{eq:varSub}, the joint density described by the graphical model can be written as
\begin{align}
	\nonumber
	p(\mathbf{o}, \mathbf{z}, \mathbf{y} | \boldsymbol{\phi}, \mathbf{G}, \boldsymbol{\sigma})
	&= p(\mathbf{o} | \mathbf{G})p(\mathbf{y} | \boldsymbol{\sigma}, \mathbf{o}, \mathbf{z})p(\mathbf{z} | \boldsymbol{\phi})\\
	\nonumber
	&= \mathcal{N}(\mathbf{o} | \mathbf{0} , \tilde{\boldsymbol{\Omega}})
	\mathcal{N}(\mathbf{y} | \mathbf{z} +\mathbf{o} , \mathbf{T})
	\mathcal{N}(\mathbf{z} | \mathbf{0}, \mathbf{C}_{\boldsymbol{\phi}})\\
	\nonumber
	&= \mathcal{N}(\mathbf{o} | \mathbf{0} , \tilde{\boldsymbol{\Omega}})
	\mathcal{N}(\mathbf{o} | \mathbf{y} -\mathbf{z} , \mathbf{T})
	\mathcal{N}(\mathbf{z} | \mathbf{0}, \mathbf{C}_{\boldsymbol{\phi}})\\
	\nonumber
	&= \mathcal{N}(\mathbf{o} | \mathbf{m}, \mathbf{C})
	\mathcal{N}(\mathbf{y} - \mathbf{z} | \mathbf{0}, \tilde{\boldsymbol{\Omega}} + \mathbf{T})
	\mathcal{N}(\mathbf{z} | \mathbf{0}, \mathbf{C}_{\boldsymbol{\phi}})\\
	\nonumber
	&= \mathcal{N}(\mathbf{o} | \mathbf{m}, \mathbf{C})
	\mathcal{N}(\mathbf{z} | \mathbf{y}, \tilde{\boldsymbol{\Omega}} + \mathbf{T})
	\mathcal{N}(\mathbf{z} | \mathbf{0}, \mathbf{C}_{\boldsymbol{\phi}})\\
	&= \mathcal{N}(\mathbf{o} | \mathbf{m}, \mathbf{C})
	\mathcal{N}(\mathbf{y} | \mathbf{0}, \tilde{\boldsymbol{\Omega}} + \mathbf{T} + \mathbf{C}_{\boldsymbol{\phi}})
	\mathcal{N}(\mathbf{z} | \boldsymbol{\mu}_z, \boldsymbol{\Sigma}_z),
\end{align}
where
\begin{align}
	\boldsymbol{\mu}_z &= \boldsymbol{\Sigma}_z (\tilde{\boldsymbol{\Omega}} + \mathbf{T})^{-1} \mathbf{y}\\
	\nonumber
	\boldsymbol{\Sigma}_z &= ((\tilde{\boldsymbol{\Omega}} + \mathbf{T})^{-1} + \mathbf{C}_{\boldsymbol{\phi}}^{-1})^{-1}\\
	\nonumber
	&= \mathbf{C}_{\boldsymbol{\phi}} - \mathbf{C}_{\boldsymbol{\phi}}(\tilde{\boldsymbol{\Omega}} + \mathbf{T} + \mathbf{C}_{\boldsymbol{\phi}})^{-1}\mathbf{C}_{\boldsymbol{\phi}}\\
	\nonumber
	&= (\tilde{\boldsymbol{\Omega}} + \mathbf{T})(\tilde{\boldsymbol{\Omega}} + \mathbf{T} + \mathbf{C}_{\boldsymbol{\phi}})^{-1}\mathbf{C}_{\boldsymbol{\phi}}\\
	&=\mathbf{C}_{\boldsymbol{\phi}}(\tilde{\boldsymbol{\Omega}} + \mathbf{T} + \mathbf{C}_{\boldsymbol{\phi}})^{-1}(\tilde{\boldsymbol{\Omega}} + \mathbf{T}).
\end{align}

After marginalizing out $\mathbf{o}$ and dividing by the marginal of $\mathbf{y}$, we get the conditional distribution
\begin{equation}
p(\mathbf{z} | \mathbf{y}, \boldsymbol{\phi}, \mathbf{G}, \boldsymbol{\sigma}) = \mathcal{N}(\mathbf{z} | \boldsymbol{\mu}_z, \boldsymbol{\Sigma}_z).
\end{equation}

%%%%%%%%%%%%%%%%%%%%%%%%%%%%%%%%%%%%%%%%%%%%%%%%%%%%%%%%%%%%%%%%%%%%%%%%%%%%%%%
%%%%%%%%%%%%%%%%%%%%%%%%%%%%%%%%%%%%%%%%%%%%%%%%%%%%%%%%%%%%%%%%%%%%%%%%%%%%%%%

\end{document}